\documentclass[conference]{IEEEtran}
\IEEEoverridecommandlockouts
\usepackage{cite}
\usepackage{amsmath,amssymb,amsfonts}
\usepackage{algorithmic}
\usepackage{graphicx}
\usepackage{textcomp}
\usepackage{pgfplotstable}
\usepackage{xcolor}
\def\BibTeX{{\rm B\kern-.05em{\sc i\kern-.025em b}\kern-.08em
    T\kern-.1667em\lower.7ex\hbox{E}\kern-.125emX}}

\IEEEpubid{\makebox[\columnwidth]{979-8-3315-2480-7/25/\$31.00~\copyright2025 IEEE \hfill}
\hspace{\columnsep}\makebox[\columnwidth]{ }}

\begin{document}

\title{LLM-Powered Knowledge Graphs for Enterprise Intelligence and Analytics\\
}

\author{
\IEEEauthorblockN{1\textsuperscript{st} Rajeev Kumar}
\IEEEauthorblockA{\textit{Gen AI Research} \\
\textit{Althire AI}\\
San Francisco, USA \\
rajeev@althire.ai}
\and
\IEEEauthorblockN{2\textsuperscript{nd} Kumar Ishan}
\IEEEauthorblockA{\textit{Gen AI Research} \\
\textit{Althire AI}\\
Bangaluru, India \\
ishan@althire.ai}
\and
\IEEEauthorblockN{3\textsuperscript{rd} Harishankar Kumar}
\IEEEauthorblockA{\textit{Gen AI Research} \\
\textit{Althire AI}\\
Gurgaon, India \\
hsk@althire.ai}

\and
\IEEEauthorblockN{4\textsuperscript{th} Abhinandan Singla}
\IEEEauthorblockA{\textit{Gen AI Research} \\
\textit{Althire AI}\\
Gurgaon, India \\
abhinandan@althire.ai}
}

\maketitle
\IEEEpubidadjcol

\begin{abstract}
Disconnected data silos within enterprises obstruct the extraction of actionable insights, diminishing efficiency in areas such as product development, client engagement, meeting preparation, and analytics-driven decision-making. This paper introduces a framework that uses large language models (LLMs) to unify various data sources into a comprehensive, activity-centric knowledge graph. The framework automates tasks such as entity extraction, relationship inference, and semantic enrichment, enabling advanced querying, reasoning, and analytics across data types like emails, calendars, chats, documents, and logs. Designed for enterprise flexibility, it supports applications such as contextual search, task prioritization, expertise discovery, personalized recommendations, and advanced analytics to identify trends and actionable insights. Experimental results demonstrate its success in the discovery of expertise, task management, and data-driven decision making. By integrating LLMs with knowledge graphs, this solution bridges disconnected systems and delivers intelligent analytics-powered enterprise tools.
\end{abstract}

\begin{IEEEkeywords}
Knowledge graph, Entity extraction, Relation extraction, LLM, Activity Graph, Enterprise Intelligence.
\end{IEEEkeywords}

\section{Introduction}
Today, companies face the challenge of disconnected data ecosystems, where critical information resides in silos such as emails, calendars, documents, activity logs, and other repositories. Although these data sets are rich in content, the absence of a unified semantic representation hinders meaningful insights and reduces efficiency in workflows such as task prioritization, client engagement, meeting preparation, and analytics-driven decision-making. Lack of integration limits the ability to derive trends, perform predictive analytics, or uncover actionable insights essential for informed decision-making and strategic planning. Unified knowledge graphs have emerged as a promising solution, offering the ability to model relationships between disparate data facets and allowing cohesive reasoning and representation [1],[2]. However, existing approaches often depend on rigid ontologies and system-specific implementations, making them difficult to scale and adapt to the diverse and dynamic needs of modern enterprises.

The primary motivation behind this work is to unify disparate data silos to improve enterprise workflows and decision-making.
We introduce a single, user-centric knowledge graph that integrates multiple sources (emails, calendars, chats, documents, and logs) through LLM-driven entity extraction, relationship inference, and semantic enrichment.
Our system-agnostic design avoids rigid ontologies, offering enhanced analytics for tasks such as expertise discovery, task prioritization, and anomaly detection, all while maintaining scalability and flexibility across diverse enterprise contexts.

This paper presents a novel framework for constructing and querying LLM-powered user-centric activity knowledge graphs. LLMs excel in semantic enrichment, entity extraction, and contextual reasoning, making them ideal for creating dynamic and adaptive graph structures [2]. By integrating diverse data sources, including emails, calendars, chats, logs, and documents, the framework creates a unified representation centered on user activities and organizational objectives. Beyond traditional applications, the framework supports advanced analytics by allowing the discovery of trends, patterns, and relationships that would otherwise remain hidden. This capability is particularly valuable for predictive decision making, anomaly detection, and workflow optimization. Applications range from contextual search, task prioritization, and expertise discovery to personalized recommendations and advanced data analytics tailored to organizational needs [3]-[5]. Experimental results demonstrate its effectiveness in improving workflows such as aggregating daily priorities, preparing for meetings, and conducting analytics-driven assessments, while also seamlessly adapting to new data and domains [4],[6].

This framework employs LLMs to dynamically infer relationships, enriching knowledge graphs with unrecognized connections and analytical context. It supports multimodal data integration—including textual, temporal, and behavioral data—allowing comprehensive reasoning, trend analysis, and advanced query capabilities. The user-centric design of the framework further ensures accessibility through natural language interfaces. For example, questions such as ‘‘What are my key priorities this week?’’ or ‘‘What trends are emerging in client engagement?’’ produce aggregated insights from disparate systems, providing intuitive solutions to complex enterprise challenges. By bridging the gap between fragmented data silos and analytics-powered enterprise intelligence, this work offers a scalable and adaptive approach for modern organizations to extract actionable insights and improve decision-making.

\section{Literature Review}

The integration of Large Language Models (LLMs) and knowledge graphs has gained significant attention to address fragmented enterprise data silos and enhance intelligent decision making. Knowledge graphs have long been recognized for their ability to represent complex relationships in diverse data domains, allowing contextual reasoning and semantic enrichment [2]. However, traditional approaches often rely on static ontologies, limiting adaptability to dynamic enterprise workflows. Recent advances in LLM, particularly their capabilities in entity extraction, relation inference, and contextual understanding, have further expanded the potential of knowledge graphs [1]-[2].

Several studies highlight the application of LLMs in enterprise settings, such as task prioritization, analytics-driven insights, and expertise discovery [3]. These works emphasize the value of combining LLM with retrieval-augmented generation (RAG) techniques to enhance precision in contextual retrieval and entity-relationship extraction. In addition, applications such as contextual search and task alignment have demonstrated the ability to improve productivity and decision making by bridging data silos [6]. Although prior efforts have largely focused on specific domains or static datasets, this paper extends the literature by presenting a dynamic, system-agnostic framework capable of adapting to evolving enterprise needs, providing actionable insights, and addressing real-world challenges such as meeting preparation, task management, and expertise identification.

Despite these promising directions, there are notable challenges and limitations when integrating LLM and knowledge graphs in enterprise contexts. One key concern is hallucination, where LLMs can generate inaccurate facts or relationships, necessitating robust validation mechanisms. Data privacy and security are also of great importance, particularly when enterprise data is sensitive or subject to regulatory constraints. The computational overhead of running LLM-based extraction at scale can be prohibitive, requiring careful optimization and hardware considerations. Finally, ontology mismatch can arise when merging different knowledge sources or when domain-specific ontologies conflict, highlighting the need for adaptive schema alignment and continuous ontology evolution [7]-[9]. These challenges underscore the importance of designing frameworks that not only leverage the strengths of LLMs, but also implement safeguards to maintain data integrity, scalability, and compliance.

\section{Methodology}

Our proposed framework is designed to unify multifaceted data into a single knowledge graph, seamlessly connecting information from emails, meetings, tasks, documents, and other sources. It is built with several key objectives in mind: system agnosticism, scalability, extensibility, and intelligent query processing. The framework achieves these objectives through a robust architecture that enables data ingestion, graph construction, efficient storage, and natural language querying. Using large language models (LLMs), the framework automates entity extraction, relationship inference, and contextual enrichment, ensuring a dynamic and adaptive knowledge graph structure that scales across domains. 

At the core of the framework lies a unified graph representation, where nodes represent entities such as people, topics, or events, and edges represent relationships inferred by LLMs. This graph structure dynamically updates as new data is ingested, making it adaptable to evolving data environments. Inspired by advances in the integration of LLMs with knowledge graphs [6]-[15], the framework employs five primary components: a data ingestion layer, a graph construction module, a distributed graph store, a query interface and scenario-specific extensions. Fig. 1 explains the top-level architecture of this framework. 

\begin{figure}[htbp]
    \centering
    \includegraphics[width=\columnwidth]{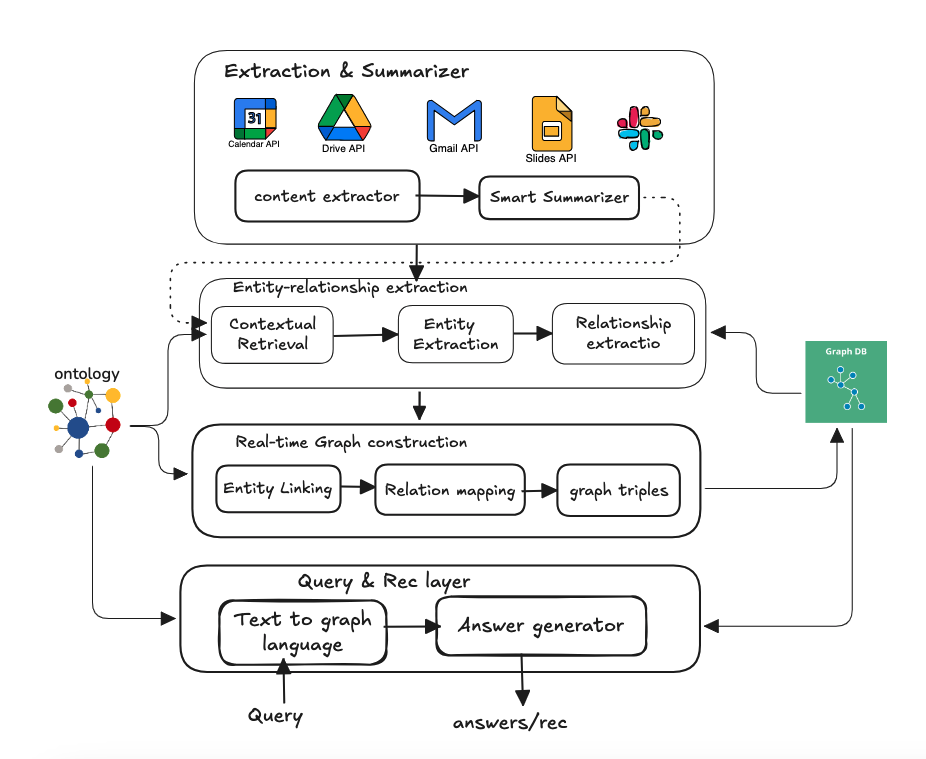}
    \caption{Framework overview for unified knowledge graph}
    \label{fig:example}
\end{figure}

\subsection{Dataset Explanation}

The data set for this experimentation was collected from consulting companies operating in various domains, including power, medicine, finance, gaming, and more, which encompasses a wide range of expertise and services. These companies not only provide consulting, but also engage in product development, covering nearly all major industries. Their operations are heavily driven by client meetings, task prioritization, and identifying the right personnel to execute specific projects, often under the challenge of high employee turnover rates.

Data were anonymized by removing identifiable information and sensitive content, ensuring compliance with privacy standards. All data were voluntarily shared by employees, contributing to a rich dataset of more than 3 million activities generated over the last two years. This data set provides a comprehensive view of consulting workflows, including meetings, task management, and resource allocation, making it ideal for studying patterns and improving operational efficiency.

\subsection{Extraction And Summarizer}

This layer collects raw data from sources such as emails, calendars, documents, customer interactions, activity logs, social media feeds, and public knowledge bases. Data retrieval is managed through APIs or crawlers, with secure permissions and access tokens. Modular extensions allow plugins for specific data types, such as email synchronization, calendar events, and document indexing, to operate at configurable intervals with opt-in or opt-out controls. To address data sensitivity, especially for emails and documents, the framework ensures secure operations in private cloud environments. Only summary-level metadata essential for graph construction is extracted, balancing robust data integration with privacy compliance.

\subsubsection{Content Extractor}

The Context Extractor is a preprocessing component that gathers all relevant textual and visual content from emails, meetings, tasks, and communications to prepare data for the SmartSummarizer. Extracts and consolidates information such as email bodies, attachments, calendar descriptions, task details, and associated images, ensuring that all contextual elements required to generate meaningful summaries are captured. The component filters unnecessary or redundant data and focuses on collecting only the input content relevant to preserving entities, activities, and relationships. This ensures that the SmartSummarizer runs efficiently with complete and high-quality contextual input.

\subsubsection{Smart-Summarizer}
The Smart-Summarizer is an LLM-powered module designed to generate concise, structured summaries while ensuring that all relevant entities, activities, and relationships are preserved in the summary text. Rather than explicitly extracting entities or relationships, the module focuses on maintaining their integrity within the generated summary, ensuring contextual completeness and relevance. It also filters out sensitive information, such as unique identifiers or private details, to ensure downstream components process only de-identified and privacy-compliant text. This approach balances rich content representation with security and compliance requirements, enabling seamless downstream integration. Fig. 2. Explains with an example.

\begin{figure}[htbp]
    \centering
    \includegraphics[width=\columnwidth]{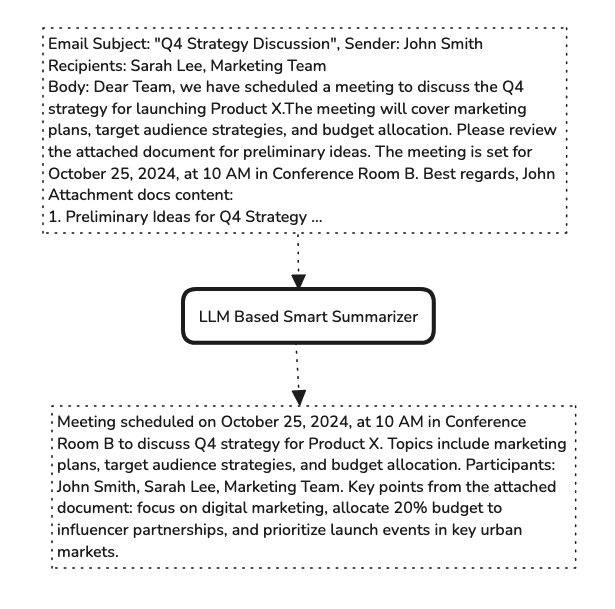}
    \caption{LLM-powered Smart-Summarizer converting input text, attached documents, and images into concise, structured summaries.}
    \label{fig:example}
\end{figure}

For example, the Smart-Summarizer processes email datasets to extract key details such as the sender, recipient, subject, and timestamps, along with embedded information like event specifics, hotel bookings, or flight details. Similarly, meeting invitations or document attachments are summarized to capture topics and relationships, including references from external links or associated files. Calendar entries are analyzed to extract participants, locations, and schedules, while LLMs infer connections, such as associating an email referencing "Project A" with a related calendar meeting. Once the summaries are enriched with these details, the data is refined and prepared for integration into the graph construction module.
\subsection{Entity-Relationship Extraction}

The Smart-Summarizer output is processed by this layer, which performs Named Entity Extraction (e.g., names, locations, dates) and Relation Extraction (e.g., traveling\_on, staying\_at, attending\_event, participating\_in). This component identifies key entities and relationships, using a contextual relevance module to retrieve pertinent information about entities from the existing knowledge graph (KG) store. This enriched context enhances the precision of LLMs based extraction. The workflow begins with contextual retrieval, followed by entity extraction and relationship extraction, ensuring an accurate mapping of interactions. At this stage, entities are not yet linked to real-world counterparts, and relationships remain unnormalized with respect to the ontology defined in the framework.

\subsubsection{Contextual Retrieval}
he Contextual Retrieval Module (CRM) employs Retrieval-Augmented Generation (RAG) techniques to enhance a given summary by retrieving additional information about related entities and their relationships from the Knowledge Graph (KG). By integrating relevant entity summaries and associated relations, the CRM provides enriched context to ensure greater precision in downstream entity and relationship extraction tasks. This added context enables a more comprehensive understanding of interactions and supports better decision making. Fig. 3. is an example of contextual information extracted for summarized text from KG through RAG. 

\begin{figure}[htbp]
    \centering
    \includegraphics[width=\columnwidth]{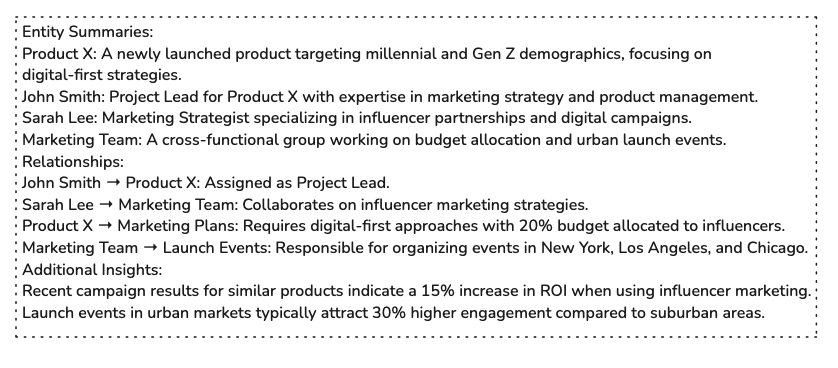}
    \caption{Contextual Retrieval Module Extracting Entity and Relationship Insights from the Knowledge Graph}
    \label{fig:example}
\end{figure}

\subsubsection{Entity Extraction}
The entity extraction process is a critical step in converting summarized content, enriched with additional contextual information, into structured data. This component leverages Large Language Models (LLMs) with carefully designed prompt engineering techniques, augmented by contextual data retrieved from the Contextual Retrieval Module (CRM). Using in-context learning, the process improves the precision and consistency of entity extraction, effectively resolving ambiguities and improving downstream workflows.

Experiments without contextual enrichment, such as using basic prompts or a few-shot examples, demonstrated suboptimal results compared to the use of enriched contextual information. The integration of CRM-provided context significantly outperformed other methods, yielding more accurate and reliable entity extraction. This approach ensures a comprehensive list of extracted entities, providing a solid foundation for subsequent tasks.

\subsubsection{Relation Extraction}
The relationship extraction process is a vital step in structuring the interactions between entities identified in the summarized content, further enriched with contextual information. This component utilizes LLMs with advanced prompt engineering techniques, incorporating both contextual data retrieved from the CRM and the extracted entities as input. Using these enriched inputs, the process enhances the precision and relevance of relationship extraction, effectively resolving ambiguities and providing meaningful connections between entities.

\subsection{Real-time Graph Construction}

The graph construction module is responsible for resolving entity ambiguities and assigning unique identifiers to maintain consistency across the knowledge graph. Recognized entities are matched to their existing identifiers, while new identifiers are assigned to previously unknown entities. Relationships and attributes are aligned with a pre-existing ontology schema, or a new schema is generated when necessary. All data is normalized to adhere to the specified ontology format, ensuring semantic compatibility throughout the graph. Each identified entity is represented as a node, with relationships inferred by the LLMs depicted as edges. For instance, a calendar meeting is represented as a meeting node connected to participant nodes, with edges labeled with properties such as "attends" or "organizes." Contextual enrichment is incorporated, enabling LLMs to infer additional relationships, such as connecting a project mentioned in a document to a related task, resulting in a more connected and enriched data structure.

\begin{figure}[htbp]
    \centering
    \includegraphics[width=\columnwidth]{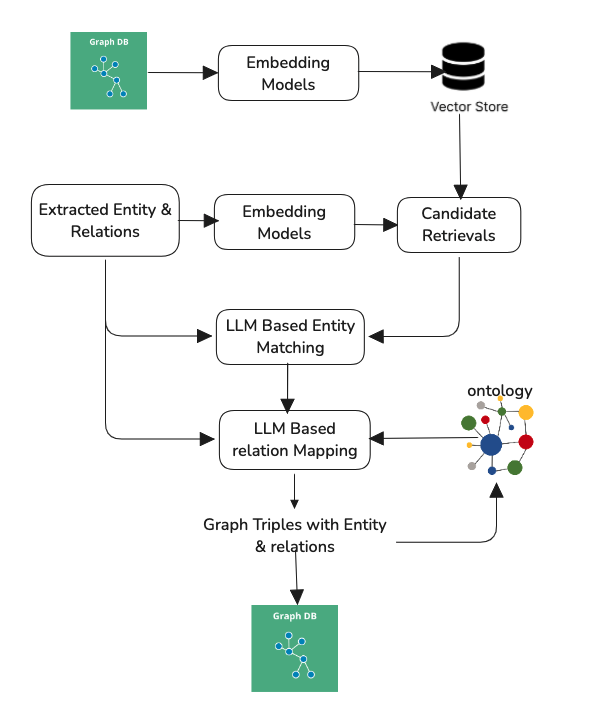}
    \caption{ Leveraging Embedding Models, Contextual Retrieval, and LLM-Based Mapping to Construct Graph Triples for Knowledge Graphs.}
    \label{fig:example}
\end{figure}

As illustrated in Fig. 4, this process begins with transforming extracted entities and relationships into vector embeddings via embedding models. Simultaneously, existing entities and relationships within the graph database undergo the same embedding process, with their vector representations stored in a vector store for efficient retrieval. The vector store facilitates the retrieval of relevant entities and relationships from the knowledge graph that align with the extracted data. The extracted candidates and entities are processed by the LLM-based entity matching module, which resolves ambiguities and accurately maps the extracted entities to existing graph entities. Following this, the LLM-based relationship mapping module integrates the matched entities with the ontology schema to infer and normalize relationships, ensuring that they align with domain-specific semantics. The resulting output is a set of graph triples representing entities, relationships, and associated contextual information, which are stored in the graph database for subsequent queries and reasoning tasks.

\subsection{Recommendations and Analytics Layer}

The Recommendations and Analytics layer combines knowledge graph data with LLM-based reasoning to provide actionable insights and analytics. Designed to meet diverse enterprise needs, this layer supports applications such as meeting preparation, task prioritization, expertise identification, and analytics-driven decision making. Customized LLM prompts with contextual learning are tailored for each scenario to achieve optimal results. Although the following sections detail three key applications, the system is currently being tested for more than five use cases, highlighting its versatility and potential for broader implementation.

\subsubsection{Expertise Discovery}
This layer enables users to identify employees with expertise in specific skills, concepts, or topics by analyzing their contributions to tasks, meetings, and communications. It is particularly valuable for fostering collaboration and streamlining task delegation. Users can upload entire discussion threads or related documents to the system, which processes the input to generate a list of domain experts. Leveraging LLMs, the system extracts and ranks skills or concepts connected to the input text, performs graph traversal on the knowledge graph, and refines results using LLM-based re-ranking to ensure precision and relevance.

For example, when a user queries "Who is the best person to consult about influencer marketing strategies?" the system navigates the knowledge graph to identify individuals linked to "influencer marketing" through documented skills, completed projects, or participation in relevant meetings. The ranked results might highlight individuals like Sarah Lee, a marketing strategist with notable contributions to recent campaigns. Similarly, for conversation-based queries like "Who has been involved in discussions about AI model optimization?" the system analyzes conversation logs, extracts contextual details, and cross-references the knowledge graph to identify and rank experts. This functionality reduces the time it takes to locate appropriate resources and promotes more effective collaboration between teams.

\subsubsection{Task Prioritization}
In this system, the knowledge graph is constructed with the user as a central node, enabling the system to understand all user-specific activities. By traversing the graph, the system provides daily or weekly recommendations on which tasks should be prioritized. In addition, it displays relevant materials, conversations, or contextual information that is needed to perform those tasks efficiently. The dynamic nature of this system ensures that any changes in priorities, such as leadership-level decisions, are immediately reflected, often before managers have the opportunity to communicate these changes. This capability significantly improves employees' ability to stay up-to-date and focus on high-priority tasks, streamlining the prioritization process, and improving overall productivity.

\subsubsection{Insights and Decision-Making}

The system leverages graph analytics and LLMs to process natural language queries, translating them into graph traversal and analytics operations. Retrieval of relevant statistics from the knowledge graph, segmentation of data as needed, and refinement of results for clarity and precision. For instance, a query like "What percentage of tasks were completed on time last quarter?" would retrieve task-completion rates, segmented by departments. The response might state: "In Q3, 75\% of the tasks were completed on time, with the marketing team achieving 85\% and the engineering team achieving 65\%.

This functionality provides decision makers with actionable insights tailored to their needs, facilitating data-driven decisions that improve performance and address inefficiencies. By providing detailed analytics in an accessible format, the system enables organizations to optimize workflows, identify bottlenecks, and align strategies with operational goals.

\subsection{System Evaluation}
We conducted a 6-month pilot study in two large companies in the finance and healthcare sectors to validate the performance of the system in real world environments. During this deployment period, our framework integrated diverse data sources, such as emails, internal chats, and project documents, while supporting workflows such as expertise discovery and task prioritization. To quantitatively assess performance, we used a combination of metrics tailored to specific functionalities, including NDCG (Normalized Discounted Cumulative Gain) for rank-based tasks and additional indicators such as user satisfaction rates, precision and recall. This approach allowed us to capture both short-term and long-term impacts of the system. The feedback of the employees highlighted benefits such as centralized knowledge access and improved task alignment, underscoring the utility and scalability of the framework in daily operations.

For expertise discovery, we measured the system's expert ranking using NDCG, supplemented by user feedback to validate recommendations. Additional metrics like Mean Reciprocal Rank (MRR) helped gauge the precision of the top-ranked results. For task prioritization, the system recommended tasks based on importance, urgency, and dependencies, and we tracked user actions (e.g., task completions) to infer implicit relevance signals. The precision at k (P @ k), the recall, and NDCG were then used to assess how effectively the system prioritized critical tasks. In analytics queries, user satisfaction served as the key metric: For example, when asked 'What percentage of tasks were completed on time last quarter?', the system retrieved statistics from the knowledge graph, refined them via LLM, and provided actionable insights.

In addition, the performance of different LLM models was assessed for the extraction of entities and relationships, as well as application-level services. Comparative experiments demonstrated that integrating contextual data from the knowledge graph improved extraction accuracy and downstream task performance. This multifaceted evaluation framework ensured a comprehensive understanding of the system's capabilities and its impact on enterprise workflows.
\section{Results AND Discussion}
The evaluation results, summarized in Tables I, II, and III, demonstrate the effectiveness of the system in various functionalities, with metrics such as precision, recall, NDCG, and satisfaction rates that highlight its impact on enterprise applications. In addition to scenario evaluations, internal components were evaluated, with entity extraction achieving 92\% accuracy and relationship extraction reaching 89\%, supported by contextual enrichment that improved task alignment by 15\%. During a six-month period, the system achieved a user adoption rate 78\% across multiple departments and successfully addressed five of the six targeted scenarios, including expertise discovery, task prioritization, and analytics. These results underscore the system's ability to enhance workflows and deliver actionable insights.
\begin{table}[ht]
    \caption{Expertise Discovery performance}
    \centering
    \renewcommand{\arraystretch}{1.5} 
    \setlength{\tabcolsep}{4pt}       
    \begin{tabular}{|p{0.15\columnwidth}|p{0.5\columnwidth}|p{0.25\columnwidth}|}
        \hline
        \textbf{Metric} & \textbf{Description} & \textbf{Performance} \\ \hline
        NDCG & Evaluates the ranking quality of suggested experts based on relevance scores from follow-up feedback. & NDCG@5=0.80, NDCG@3=0.63 \\ \hline
        MPR & Measures how quickly the most relevant expert appears in the ranked list & MPR=0.83\\ \hline
        Precision at K & Assesses the relevance of the top-k recommended experts & P@3=0.62, P@5=0.83. \\ \hline
    \end{tabular}
    \label{tab:single_column}
\end{table}

\begin{table}[ht]
    \caption{Task prioritization performance}
    \centering
    \renewcommand{\arraystretch}{1.5} 
    \setlength{\tabcolsep}{4pt}       
    \begin{tabular}{|p{0.15\columnwidth}|p{0.5\columnwidth}|p{0.25\columnwidth}|}
        \hline
        \textbf{Metric} & \textbf{Description} & \textbf{Performance} \\ \hline
        NDCG & Evaluates the prioritzation of tasks based on user implicit feedback.& NDCG@5=0.72, NDCG@3=0.59 \\ \hline
        Precision at K & Measures how accurately top-k recommended tasks align with completed tasks & P@3=0.57, P@5=0.80. \\ \hline
        Recall & Measures how many critical tasks were included in recommendations. & Recall=0.83. \\ \hline
    \end{tabular}
    \label{tab:single_column}
\end{table}

\begin{table}[ht]
    \caption{Analytical Queries performance}
    \centering
    \renewcommand{\arraystretch}{1.5} 
    \setlength{\tabcolsep}{4pt}       
    \begin{tabular}{|p{0.15\columnwidth}|p{0.5\columnwidth}|p{0.25\columnwidth}|}
        \hline
        \textbf{Metric} & \textbf{Description} & \textbf{Performance} \\ \hline
        Satisfaction Rate & Assesses user satisfaction with analytical insights delivered by the system.& 83\% positive feedback \\ \hline
        Accuracy & Compares system-generated analytics to manual calculations. & Accuracy = 0.86. \\ \hline
    \end{tabular}
    \label{tab:single_column}
\end{table}

The results highlight the effectiveness of integrating LLM with knowledge graphs in addressing enterprise challenges, demonstrating significant improvements in task prioritization, expertise discovery, and analytics-driven decision making. 

\section{Conclusions}

This paper presents a novel framework that integrates Large Language Models (LLMs) with knowledge graphs to address key enterprise challenges, including expertise discovery, task prioritization, and analytics-driven decision-making. Using contextual retrieval and advanced entity-relationship extraction techniques, the system dynamically generates actionable insights, personalized recommendations, and precise analytics. The evaluation results demonstrate high performance in metrics such as NDCG, precision, recall, and user satisfaction, with notable improvements in prioritization accuracy and expert identification. This framework improves productivity, collaboration, and decision-making in enterprise environments, bridges fragmented data silos, and enables intelligent, scalable solutions tailored to dynamic organizational needs. Future work will expand the system's applications and further refine its scalability and adaptability. 

\section{Future Work}
In the future, we plan to enrich the knowledge graph with multimodal data, including images and audio, to provide more comprehensive visual and auditory signals. We will also fine-tune LLM models for specific tasks to achieve higher performance metrics and explore real-time collaboration features, such as linking internal entities to external knowledge graphs (e.g., Wikipedia) for deeper insights. This could help organizations understand the impact of external events on task priorities or identify large-scale opportunities. Furthermore, scaling the system for larger datasets and incorporating advanced predictive analytics will enhance decision making by forecasting potential risks, recommending proactive measures, or automatically spotting emerging trends. Finally, expanding the framework to domains such as risk assessment, compliance monitoring, and personalized employee training will broaden its utility and impact in diverse enterprise workflows.


\begin{thebibliography}{00}
\bibitem{b1} Bommasani, R., Hudson, D., Adcock, A., et al. (2021). On the opportunities and risks of foundation models.
\bibitem{b2} Ji, S., Pan, S., Cambria, E., et al. (2022). A survey on knowledge graphs: Representation, construction, and applications. IEEE Transactions on Neural Networks and Learning Systems.
\bibitem{b3} Guo, Y., Gao, H., Li, J., Pan, J. (2022). Conversational query graphs for intuitive knowledge navigation. International Journal of Semantic Computing.
\bibitem{b4} Wang, Q., Mao, Z., Wang, B., Guo, L. (2022). Knowledge graph embedding: A survey of approaches and applications. IEEE Transactions on Big Data.
\bibitem{b5} Zhang, Y., Liu, J., Wang, F., et al. (2021). Task prioritization in multi-faceted knowledge graphs. In Proceedings of the 27th ACM SIGKDD Conference.
\bibitem{b6} Ibrahim, N., Aboulela, S., Ibrahim, A., Kashef, R. (2024). A survey on augmenting knowledge graphs (KGs) with large language models (LLMs): Models, evaluation metrics, benchmarks, and challenges. Discover Artificial Intelligence, 4(76). Springer. DOI: 10.1007/s44163-024-00175-8.
\bibitem{b7} Pan, S., Luo, L., Wang, Y., Chen, C., Wang, J., Wu, X. (2024). Unifying large language models and knowledge graphs: A roadmap. arXiv preprint arXiv:2306.08302.
\bibitem{b8} Jin, B., Liu, G., Han, C., Jiang, M., Ji, H., Han, J. (2024). Large language models on graphs: A comprehensive survey. arXiv preprint arXiv:2312.02783.
\bibitem{b9} Agrawal, G., Kumarage, T., Alghamdi, Z., Liu, H. (2023). Can knowledge graphs reduce hallucinations in LLMs? A survey. arXiv preprint arXiv:2311.07914.
\bibitem{b10} Khorashadizadeh, H., Amara, F. Z., Ezzabady, M., Ieng, F., Tiwari, S., Mihindukulasooriya, N., Groppe, J., Sahri, S., Benamara, F., Groppe, S. (2024). Research trends for the interplay between large language models and knowledge graphs. arXiv preprint arXiv:2406.08223.
\bibitem{b11} Chen, Z., Zhang, N., Chen, H. (2024). Knowledge graphs meet multi-modal learning: A comprehensive survey. arXiv preprint arXiv:2402.05391.
\bibitem{b12} Li, H., Appleby, G., Suh, A. (2024). A preliminary roadmap for LLMs as assistants in exploring, analyzing, and visualizing knowledge graphs. arXiv preprint arXiv:2404.01425.
\bibitem{b13} He, X. (2024). Awesome-Graph-LLM. GitHub Repository. https://github.com/XiaoxinHe/Awesome-Graph-LLM.
\bibitem{b14} Luo, R. (2024). Awesome-LLM-KG: Awesome papers about unifying LLMs and KGs. GitHub Repository. https://github.com/RManLuo/Awesome-LLM-KG.
\bibitem{b15} UpcomAI. (2024). KG-LLMs-papers: A repository for knowledge graph and large language model papers. GitHub Repository. https://github.com/UpcomAI/KG-LLMs-papers.
\end{thebibliography}
\end{document}